\documentclass{article}

\usepackage{PRIMEarxiv}

\usepackage[utf8]{inputenc} 
\usepackage[T1]{fontenc}    
\usepackage{hyperref}       
\usepackage{url}            
\usepackage{booktabs}       
\usepackage{amsfonts}       
\usepackage{nicefrac}       
\usepackage{microtype}      
\usepackage{lipsum}
\usepackage{fancyhdr}       
\usepackage{graphicx}       
\graphicspath{{media/}}     
\usepackage{amssymb}
\usepackage{times}
\usepackage{latexsym}
\usepackage{amsmath}
\usepackage{amsthm}
\usepackage{booktabs}
\usepackage{multirow}
\usepackage{graphicx}
\usepackage{threeparttable}

\usepackage{enumitem}
\usepackage{float}
\usepackage{subcaption}
\title{Advancing Relation Extraction through Language Probing with Exemplars from Set Co-Expansion}

\author{
  Yerong Li \\
  Department of Computer Science, University of Illinois at Urbana-Champaign, USA \\
  \texttt{yerong2@illinois.edu}
  \And
  Roxana Girju \\
  Linguistics Department, University of Illinois at Urbana-Champaign, USA \\
  \texttt{girju@illinois.edu}
}

\usepackage{bibentry}

\begin{document}

\maketitle

\begin{abstract}
Relation Extraction (RE) is a pivotal task in automatically extracting structured information from unstructured text. In this paper, we present a multi-faceted approach that integrates representative examples and through co-set expansion. The primary goal of our method is to enhance relation classification accuracy and mitigating confusion between contrastive classes.

Our approach begins by seeding each relationship class with representative examples. Subsequently, our co-set expansion algorithm enriches training objectives by incorporating similarity measures between target pairs and representative pairs from the target class. Moreover, the co-set expansion process involves a class ranking procedure that takes into account exemplars from contrastive classes. Contextual details encompassing relation mentions are harnessed via context-free Hearst patterns to ascertain contextual similarity.

Empirical evaluation demonstrates the efficacy of our co-set expansion approach, resulting in a significant enhancement of relation classification performance. Our method achieves an observed margin of at least 1 percent improvement in accuracy in most settings, on top of existing fine-tuning approaches. To further refine our approach, we conduct an in-depth analysis that focuses on tuning contrastive examples. This strategic selection and tuning effectively reduce confusion between classes sharing similarities, leading to a more precise classification process.

Experimental results underscore the effectiveness of our proposed framework for relation extraction. The synergy between co-set expansion and context-aware prompt tuning substantially contributes to improved classification accuracy. Furthermore, the reduction in confusion between contrastive classes through contrastive examples tuning validates the robustness and reliability of our method.


\end{abstract}
\section{Introduction}
Relation extraction (RE) is a fundamental process in deciphering relationships between entity mentions within a given textual context. For example, consider the sentence "Mark Fisher writes for Dayton Daily News," where an RE model would discern the relation "per:employee\_of" between the entities "Mark Fisher" and "Dayton Daily News." This task holds paramount significance in comprehending natural language and serves as a foundational step in constructing knowledge bases. The effectiveness of advanced RE models carries profound implications for a diverse array of knowledge-driven downstream tasks, including question answering \cite{xu-etal-2016-question, li-etal-2019-entity, yasunaga-etal-2021-qa}, narrative prediction\cite{Chen_Chen_Yu_2019, alt-etal-2019-fine}, and dialogue systems \cite{liu-etal-2018-knowledge, zhao-etal-2020-knowledge-grounded} etc. These collective endeavors contribute to advancing the field of relation extraction and its applicability to a wide spectrum of natural language understanding tasks etc.

With the grounding role of relation extraction in language understanding, many recent studies have increasingly framed RE as a multi-class classification task\cite{zhou-etal-2016-attention,joshi2019spanbert, yamada-etal-2020-luke, han2022ptr, han-etal-2022-generative}. Despite the demonstrated success of fine-tuning PLMs, one pivotal challenge has emerged, which is centered around the substantial incongruity between the objective forms utilized during the pre-training phase and those encountered during the fine-tuning process. This discrepancy poses a significant hindrance to fully exploiting the profound reservoir of knowledge encapsulated within PLMs. To elaborate, the pre-training phase typically employs a cloze-style task, exemplified by sequential and masked language models, for predicting target words. In contrast, the myriad of diverse downstream tasks PLMs are tuned on may manifest distinct objective forms: classification, and sequence labeling,  generation etc. Prompt tuning\cite{schick2020small, schick2020exploiting, gao-etal-2021-making} stands as a promising avenue that not only resolves the contrasting objectives but also leverages the intrinsic capabilities of PLMs by skillfully aligning cloze-style reasoning from pre-training with the requisites of classification tasks during fine-tuning. Its notable performance is evidenced by its adeptness in bridging the divide between pre-training tasks and diverse downstream tasks, underscoring its efficacy. In the domain of information extraction, the exploration of fine-tuning pre-trained language models for supervised relation extraction has yielded promising outcomes in recognizing a diverse range of relations\cite{han2022ptr,  han-etal-2022-generative}. 

Upon closer analysis of prediction errors and confusion matrices, prompt-based methods reveal a limitation: while they capture contextual relations, they struggle to fully grasp the unique characteristics of example relation pairs and their differentiation from others in one dataset. This gap underscores the need for more refined techniques in relation extraction. Despite progress in prompt tuning and pre-trained transformer models, room for improvement remains in achieving a comprehensive understanding of relation pair intricacies. Our integration of set co-expansion addresses these limitations. By seeding from a small set of exemplar pairs and incorporating similarity measures, our method enhances contextual understanding and introduces a dynamic ranking mechanism that accommodates confusing class labels. This expansion process also involves a class ranking procedure that incorporates exemplars from contrastive classes, offering a more comprehensive representation of relation pairs. These advancements empower our model with a distinct advantage in understanding and distinguishing example relation pairs, thus refining and enhancing the relation extraction process.

\section{Related Work}
\subsection{Prompt-tuning}
Prompt-tuning methods, driven by the rise of transformer-based language models, have exhibited remarkable efficacy in diverse NLP tasks. Notably, numerous studies have showcased the advancement of prompt-tuning through meticulously crafted manual prompts \cite{schick2020exploiting, schick2020small, bendavid2022pada, lester-etal-2021-power, lu-etal-2022-fantastically, le-scao-rush-2021-many}. Some approaches, such as those by \cite{hu-etal-2022-knowledgeable, chen2022knowprompt}, integrate external knowledge into the verbalizer through the implementation of calibration techniques. \cite{ding-etal-2022-prompt} extend the scope of prompt-tuning to entity typing, employing prompt-learning methodologies to construct entity-centric verbalizers and templates. In an endeavor to streamline prompt design, automatic prompt generation techniques have been explored, including the automatic generation of anchor words and templates. \cite{shin-etal-2020-autoprompt} introduce a novel gradient-guided search mechanism for the automatic generation of templates and label words from the vocabulary, while \cite{gao-etal-2021-making} incorporate the seq2seq pre-trained model T5 into the template search process. More recently, continuous prompts have been introduced \cite{hambardzumyan-etal-2021-warp, li-liang-2021-prefix, hambardzumyan-etal-2021-warp}, focusing on leveraging learnable continuous embeddings as prompt templates, departing from conventional label word methodologies. However, it is crucial to highlight that while these efforts contribute significantly to the broader prompt-tuning landscape, their direct applicability to the domain of Relation Extraction (RE) requires careful consideration.

\subsection{Set Expansion}

Corpus-based set expansion involves the identification of a comprehensive set of entities sharing the same semantic class within a given corpus, often based on a small set of initial seed entities. This task has led to a multitude of investigations in the realm of entity set expansion. Early approaches, such as EgoSet \cite{rong2016egoset} and SetExpan \cite{shen2017setexpan}, employ iterative bootstrapping techniques that leverage skip-gram features and entity ranking for expansion. Building upon these foundations, subsequent innovations like SetExpander \cite{mamou-etal-2018-term} and CaSE \cite{yu2019corpus} delve into capturing distributional similarity among words for expansion, utilizing context-free word embeddings. In the contemporary landscape, the evolution of entity set expansion has been propelled by the advent of CGExpan \cite{zhang2020empower}, which introduces language model probing to elevate the expansion process.

The ramifications of successful entity set expansion extend far beyond its immediate task, offering a potent catalyst for a plethora of downstream applications. Information extraction, taxonomy induction, question answering, and web search stand among the manifold domains that benefit from the insights garnered through entity set expansion. For instance, \cite{wang2019distantly} harness set expansion in a distantly supervised framework for named entity recognition, while \cite{shen2020taxoexpan} and \cite{lee2022taxocom} utilize set expansion to enhance self-supervised labels for taxonomy expansion. In the context of this study, we delve into set expansion for relational pairs, thereby augmenting the language model's capacity to comprehend exemplar pairs within the domain of relation classification. The synthesis of these diverse endeavors forms the backdrop against which we explore and elucidate the efficacy of set expansion methods.
\section{Proposed Methodology}
\label{sec:proposed_method}
\subsection{Constrastive Class Ranking}
Firstly, we introduce a corpus-based similarity measure to evaluate the similarity between a pair of entities and a given relation class name. The process involves constructing 12 entity probing queries by masking the target entity pair in 2 predefined patterns known as Hearst patterns. These queries are then utilized to query a pre-trained language model (LM) and obtain 12 pairs of token embeddings represented by the \texttt{[MASK]} tokens. For each contextualized representation pair, we average the two for a contextualized representation of the pair. These masked pair representations are denoted as $\mathbf{X}_c$. 

In a similar vein, we employ the technique of averaging the masked representation of the target entity pair within the query sentence, leading to the derivation of the masked representation of the pair, denoted as $\mathbf{x}_p$.
Then the similarity of the pair $p$ and class $c$\footnote{Here we pick the top-k cosine similarity scores and take the average.} The inherent alignment of contrastive class ranking with multi-class classification is evident. To derive the class score, we performed a weighted average that combines both these aspects, i.e. equation \ref{eq:sim}:
\begin{align}
\text{score}(p, c) = \frac{1}{k} \sum_{\mathbf{x}_c \in \text{top}_k^{\cos}(\mathbf{X}_c)} \cos(\mathbf{x}_p, \mathbf{x}_c)\\
s = s_\text{CLS} + \lambda \text{score}(p, c)
\label{eq:sim}
\end{align}
\begin{figure*}
  \centering
\includegraphics[width=1\textwidth]{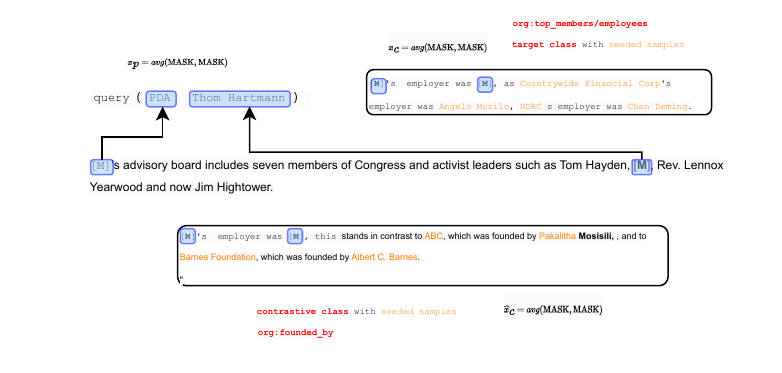}
  \caption{Context free represenation with seeded examples: In this sentence, Thom Hartmann is identified as an employee of the organization PDA, and the relation between them should be \texttt{org:top\_members/employees}. To obtain a masked representation of the pair, the contextualized representations of "PDA" and "Thom Hartmann" are averaged. Similarly, in the analogous and contrastive Hearst patterns,  we average the the contextual representations of two  \texttt{[MASK]} tokens to get a context-free representation of the paired position.}
  \label{fig:demo}
\end{figure*}

These masked representations provide a robust comparison between target entity pairs and seeded pairs within a class. They enable the evaluation of contextual embeddings obtained from the LM, facilitating the assessment of similarity between entities and relation class names. 
Based on the given definition of pair-class similarity score, we have the flexibility to select any seeded pair belonging to class $c$. By calculating its similarity score with other classes present in the corpus, we can generate a ranked list of class names that are potentially ambiguous. For a seeded candidate pair set with a size of $|\mathbf{X}_c|$, we can generate $|\mathbf{X}_c|$ ranked lists denoted as $L_1$, $L_2$, $L_3$, ..., $L_{|\mathbf{X}_c|}$. Each list corresponds to an entity pair within the set. Contrastive classes should exhibit high confusion with the positive class, specifically chosen from the top-ranked relation categories while excluding the positive class itself. These classes are identified as the most perplexing counterparts to the positive class. During the fine-tuning process, the dynamic ranking of contrasting classes evolves dynamically. As illustrated in \ref{tab:contrastive_classes}, the table offers a snapshot of the selected contrastive classes.

\begin{table*}[htbp]
\centering
\caption{A Snapshot of Contrastive Classes Selected During Tuning KnowPrompt on the reTARED Dataset}
\label{tab:contrastive_classes}
\begin{tabular}{|l|ll|}
\hline
\textbf{Positive Class} & \multicolumn{2}{l|}{\textbf{Contrastive Classes}} \\ \hline
per:charges & per:sentence & per:allegations \\
& per:indictment & per:offense \\
& no\_relation & org:top\_members/employees \\ \hline
per:date\_of\_death & per:age & per:country\_of\_birth \\
& per:origin & no\_relation \\
& per:city\_of\_death & per:date\_of\_birth \\ \hline
org:founded\_by & org:shareholders & per:other\_family \\
& org:members & org:political/religious\_affiliation \\
& org:number\_of\_employees/members & no\_relation \\ \hline
per:city\_of\_birth & per:stateorprovince\_of\_birth & no\_relation \\
& per:country\_of\_birth & per:stateorprovinces\_of\_residence \\
& per:origin & per:city\_of\_death \\ \hline
$\cdots$ &  $\cdots$  &\\ \hline
per:religion & org:political\/religious\_affiliation & per:origin \\
& per:other\_family & org:member\_of \\
& org:shareholders & per:title \\ \hline
\end{tabular}
\caption{Examples of seeded examples from ReTACRED dataset}
\label{tb:seed}
\end{table*}


\begin{table*}[h!]
\centering
\begin{tabular}{|c|c|}
\hline
\multicolumn{1}{|l|}{}     & seeds                                                                                        \\ \hline
org:top\_members/employees & \begin{tabular}[c]{@{}c@{}}('Countrywide Financial Corp', 'Angelo Mozilo')\\ ('China Charity Federation', 'Liu Guolin')\\ ('NDRC', 'Chen Deming')\\ ('National Restaurant Association', 'Mike Gibbons')\\
('Pacific Asia Travel Association','Peter de Jong')\end{tabular} \\ \hline
org:country\_of\_branch    & \begin{tabular}[c]{@{}c@{}}('NDRC', 'China')\\ ('American Association of University Women', 'United States')\\ ('National Development and Reform Commission', 'China')\\ ('Nuclear Decommissioning Authority', 'UK')\\ ('American Life Insurance company', 'US')\end{tabular} \\ \hline
per:country\_of\_death    & \begin{tabular}[c]{@{}c@{}}('his', 'Finland')\\ ('Vladimir Ladyzhenskiy', 'Finland')\\ ('Meredith Kercher', 'Italy')\\ ('Terry Jupp', 'England')\\ ('Wen Qiang', 'China')\end{tabular} \\ \hline
...    & ... \\ \hline
org:political/religious\_affiliation    & \begin{tabular}[c]{@{}c@{}}('Corporate Library', 'independent')\\ ('Organisation of Islamic Cooperation', 'Muslim')\\ ('High Point Church', 'nondenominational')\\ ('ADF', 'Muslim')\\ ('Corporate Library', 'independent')\end{tabular} \\ \hline
\end{tabular}
\end{table*}

\subsection{Exemplar Expansion}

To augment the seeded sets for each category and bolster the efficacy of relation extraction, we leverage a potent technique co-expanding exemplar pairs for contrastive relation categories. This innovative facet harnesses co-set expansion principles to effectively amplify the pool of exemplars within each category.

Our exemplar expansion process operates in harmony with the overarching goal of relation extraction. By capitalizing on co-set expansion, we amplify the representative examples for each category, thereby fostering a more comprehensive and nuanced understanding of potential relationships.

In scenarios where the co-set expansion process introduces potential inaccuracies in selecting exemplars for each category, our approach employs an ensemble algorithm to enhance the quality and robustness of entity selection. our strategy unfolds in a systematic manner, with the aim of refining entity selection. As the expansion process iterates, wrong entity pairs may be included in the set and cause semantic drifting.  Specifically, we repeatedly sample entity pairs from the current entity set $\mathbf{X}_c$, denoted as $\mathbf{X}^s_c$. Unlike the original class ranking depicted in Equation \ref{eq:sim}, where only the top-scoring item is chosen, we carry out $T$ samplings, each involving the calculation of scores from the current entity set $\mathbf{X}_c$. Similarly, we define $\text{score}^s_c$ over sampled exemplars from the exemplar set $\mathbf{X}_{c_\mathbf{N}}$ of $c_\mathbf{N}$, where $c_\mathbf{N}$ is the contrastive class of class $c$. In equation \ref{eq:sample} we use $\hat{\mathbf{C}}$ to denote the set of all constrastive class of target class $c$ in short, so $c_\mathbf{N}$ belongs to $\hat{\mathbf{C}}$\footnote{Note that $\hat{\mathbf{x}}_c$ \ref{fig:demo} from the contrastive class and $\mathbf{x}_c$ from the original class are padded differently. $\hat{\mathbf{x}}_c$ uses  contrastive patterns while $\mathbf{x}_c$ uses analogous patterns as depicted in Figure \ref{fig:demo}.}.  The average of the two scores provides the rank of the pair. This iterative process generates $T$ entity pair ranked lists. 
\begin{align}
\text{score}^s (p, c) = \frac{1}{\lvert \mathbf{X}^c_s \rvert}
\sum_{\substack{\mathbf{x}_{c} \in \mathbf{X}^s_c \\ \mathbf{X}^s_c\subseteq \mathbf{X}_c}}
\cos(\mathbf{x}_p, \mathbf{x}_{c})
\\
\text{score}^s_\mathbf{N} (p, c)= \frac{1}{\lvert \hat{\mathbf{C}} \rvert}\sum_{c_\mathbf{N} \in \hat{\mathbf{C}}}\frac{1}{\lvert \mathbf{X}^c_s \rvert}\sum_{\substack{\hat{\mathbf{x}}_{c} \in \mathbf{X}^s_c \\ \mathbf{X}^s_c\subseteq \mathbf{X}_{c_\mathbf{N}}}} \left(\cos(\mathbf{x}_p, \hat{\mathbf{x}}_{c})\right)\\
r(p, c)=\sqrt{\text{score}^s (p, c)\times\text{score}^s_\mathbf{N}(p,c)}\nonumber 
\label{eq:sample}
\end{align}
Within each respective class-ranked list, the optimal relation category, denoted as $c$, should hold a higher rank compared to any of the negative class names in the class name set $c_\mathbf{N}$. As a result, the subsequent ensemble function aggregates ranked scores from all $T$ lists.
\begin{align}
S(p, c) = \sum_{t=1}^T \Biggl(& (r(p, c) + \mathbb{I}[p \in \mathbf{X}_c])\nonumber \\\times 
& \mathbb{I}\left[ r(p, c) > \max_{c_\mathbf{N} \in \hat{C}}(r(p, c_\mathbf{N})) \right] \Biggr)
\end{align}

Crucially, the co-set expansion mechanism is designed to consider not just the target category, but also thoughtfully chosen exemplars from contrastive categories, where the selection of contrastive categories for each relation class is determined through the negative class ranking procedure described earlier. This comprehensive approach enhances the model's adaptability, ensuring its adeptness in capturing nuanced relationships within diverse contextual scenarios.

Once our exemplar expansion process is executed, the enriched exemplar pool is subjected to class-pair similarity score calculations. This assessment encapsulates both the target category and the diverse exemplars from contrastive categories. By considering this comprehensive array of entities, our model becomes adept at discerning relationships in a wider spectrum of scenarios.

The integration of exemplar expansion within our framework offers a significant enhancement to relation extraction. By harnessing co-set expansion to amplify the exemplar sets for each category, we fortify our model's ability to accurately extract relationships across diverse contexts. This innovative approach stands as a testament to our commitment to advancing the field of relation extraction.

\section{Experiments}
\subsection{Experimental Settings}
\subsubsection{Datasets}
\begin{table}[h]
\centering
\begin{tabular}{|l|c|c|c|c|}
\hline
\textbf{Dataset} & \textbf{\#train} & \textbf{\#dev} & \textbf{\#test} & \textbf{\#rel} \\
\hline
TACREV & 68,124 & 22,631 & 15,509 & 42 \\
\hline
ReTACRED & 58,465 & 19,584 & 13,418 & 40 \\
\hline
SemEval & 6,507 & 1,493 & 2,717 & 19 \\
\hline
\end{tabular}
\caption{Dataset Statistics}
\label{tb:datasetstatistics}
\end{table}
In our study, we investigate the performance of various relation classification approaches across a range of diverse datasets. Specifically, we employ three distinct datasets: TACREV \cite{alt-etal-2020-tacred}, ReTACRED \cite{stoica2021retacred}, and SemEval 2010 Task 8 \cite{hendrickx-etal-2010-semeval}.

TACREV, a dataset originating from the seminal work of \cite{alt-etal-2020-tacred}, serves as a foundational cornerstone in our analysis. Comprising a total of 42 classes, including the "no-relation" category, TACREV provides a rich and comprehensive representation of various relations. This dataset undertakes a meticulous reexamination of the original TACRED dataset, rectifying errors that may have propagated in the development and test sets. The integrity of the training set remains preserved, ensuring consistency in its utilization.

ReTACRED, an alternative rendition of the TACRED dataset as proposed by \cite{stoica2021retacred}, is characterized by its comprehensive remediation of inherent limitations. This intricate process involves a holistic reconfiguration of the training, development, and test sets, coupled with a strategic reevaluation of select relation types.

SemEval 2010 Task 8, widely recognized as SemEval, stands as a traditional yet essential benchmark for relation classification. Boasting a focal emphasis on nine discernible relations, this dataset encompasses both bidirectional relations and a distinctive "Other" category. The intricate interplay of these diverse datasets forms the backdrop against which we assess and evaluate various relation classification methods.

The detailed statistics of the datasets, including the number of training, development, and test instances, as well as the total number of unique relations, are presented in Table \ref{tb:datasetstatistics}.
\subsubsection{Baselines}
In this section, we introduce the key baselines employed in our study for relation extraction. These baselines represent a diverse set of pre-training methods that have showcased significant advancements in the field of natural language processing (NLP) and relation extraction. Each baseline offers a distinct approach to enhancing text understanding and extracting relational information from textual data.
\begin{itemize}
\item \textbf{SPANBERT}  SPANBERT\cite{joshi2019spanbert} presents a unique pre-training method that enhances text span representation and prediction. Differing from BERT, SPANBERT masks continuous token spans, using a span boundary objective (SBO) to predict masked span content solely from boundary token representations. This consistently outperforms BERT and other methods, particularly in span selection tasks like question answering and coreference resolution. SPANBERT's span-focused approach and auxiliary objective yield remarkable results in relation extraction and NLP applications.
\item \textbf{LUKE} LUKE\cite{yamada-etal-2020-luke} introduces innovative pretrained contextualized word and entity representations through a bidirectional transformer architecture. Unlike previous methods, LUKE treats words and entities as separate tokens, generating contextualized representations for both. Its pretraining task involves a masked language model applied to a large entity-annotated Wikipedia corpus. LUKE incorporates an entity-aware self-attention mechanism within the transformer, resulting in improved performance for entity-related tasks and surpassing existing approaches in empirical evaluations.
\item \textbf{PTR} PTR (Prompt Tuning with Rules) \cite{han2022ptr} leverages manually crafted sub-templates and logical rules to create extensive templates. The introduction of virtual tokens, whose embeddings are fine-tuned along with pre-trained language model parameters using training data, enhances the template's representation capabilities. This innovative combination of actual and virtual tokens within the template demonstrates substantial enhancements in relation classification tasks, underscoring the effectiveness of PTR's prompt design approach. Drawing inspiration from the relation prompts employed in PTR, we adopted a similar strategy to construct Hearst patterns.
\item \textbf{KnowPrompt} KnowPrompt\cite{chen2022knowprompt} draws inspiration from the lineage of models like SPANBERTand KNOWBERT that harness knowledge to enrich various aspects of NLP. While these models leverage external sources for relational knowledge, our focus remains on learning directly from text. We further compare with state-of-the-art methods like multi-view graph-based BERT models for relation extraction, highlighting KnowPrompt's efficacy. In our work, we center on integrating knowledge from relation labels into prompt-tuning, introducing a Knowledge-aware Prompt-tuning approach with synergistic optimization. This involves injecting latent knowledge from relation labels into prompt construction through learnable virtual type and answer words, subsequently enhancing their representation through structured constraints.
\end{itemize}

In our experiments, we fine-tune the classification header of these baselines in conjunction with co-set expansion as described in Equation \ref{eq:sim}. It's important to highlight that during the fine-tuning and class ranking stages, we select the highest-scoring entity pairs from each exemplar set. However, when dealing with exemplar examples, we opt for sampling from the exemplar set (as outlined in Equation \ref{eq:sample}), taking into consideration the presence of noise introduced during the expansion process.
\subsection{Results on Relation Classification}
\begin{table*}[h!]
\centering
\begin{tabular}{|l|c|c|c|c|}
\hline
\textbf{Model} & & \textbf{TACREV} & \textbf{ReTACRED} & \textbf{SEMEVAL} \\
\hline
\multirow{3}{*}{\textbf{SPANBERT}\cite{joshi2019spanbert}} & \textbf{original} & 77.4 & 85.6 & - \\
\cline{2-5}
& \textbf{co w/o neg} & 78.7 & 87.0 & - \\
\cline{2-5}
& \textbf{co} & 79.9 & 87.6 & - \\
\hline
\multirow{3}{*}{\textbf{LUKE}\cite{yamada-etal-2020-luke}}& \textbf{original} & 80.8 & 90.1 & - \\
\cline{2-5}
& \textbf{co w/o neg} & 80.1 & 90.7 & - \\
\cline{2-5}
& \textbf{co} & 81.5 & 90.3 & - \\
\hline
\multirow{3}{*}{\textbf{PTR}\cite{han2022ptr}} & \textbf{original} & 81.1 & 90.4 & 88.7 \\
\cline{2-5}
& \textbf{co w/o neg} & 84.0 & 89.9 & 89.2 \\
\cline{2-5}
& \textbf{co}& \textbf{84.76} & 92.2& 90.4  \\
\hline
\multirow{3}{*}{\textbf{KnowPrompt}\cite{chen2022knowprompt}} & \textbf{original} &  82.4 & 90.6 &  90.0 \\
\cline{2-5}
& \textbf{co w/o neg} &  83.7 & 90.5 & 90.8 \\
\cline{2-5}
& \textbf{co} & 83.4 & \textbf{92.4} & \textbf{91.6} \\
\hline
\end{tabular}

\caption{Comparative Performance Analysis of Relation Extraction Models on TACREV, ReTACRED, and SEMEVAL Datasets. The \textbf{original} category represents the baseline performance, \textbf{co w/o neg} signifies co-set expansion with contrastive scores from exemplars of contrastive classes, while \textbf{co} encompasses full co-set expansion with contrastive comparisons.}
\label{tb:results}
\end{table*}
\begin{figure*}[ht]
  \centering
  \begin{subfigure}[b]{0.45\textwidth}
    \includegraphics[width=\textwidth]{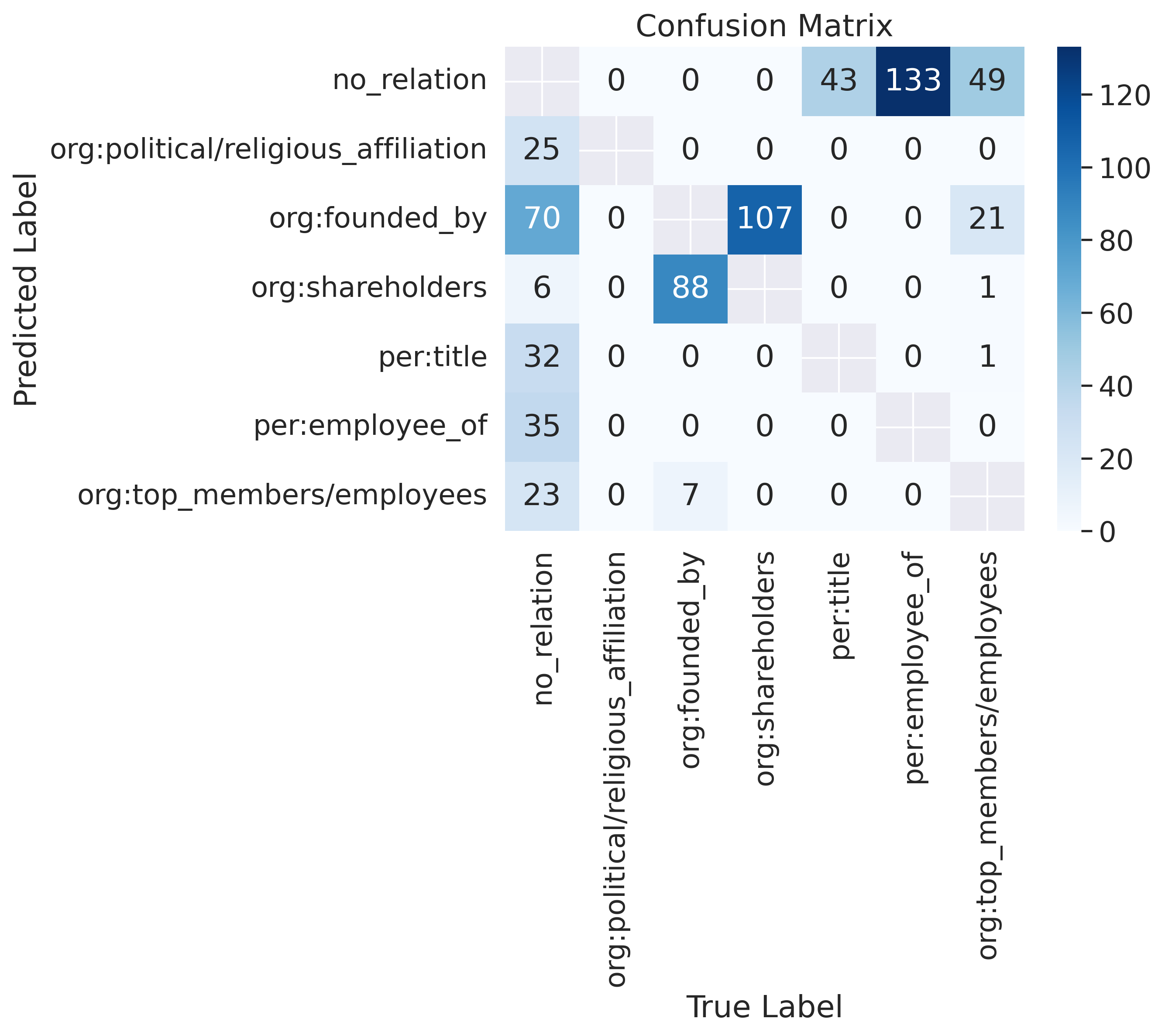}
    \caption{Confusion Matrix w/o Set Co-Expansion}
  \end{subfigure}%
  \hfill
  \begin{subfigure}[b]{0.45\textwidth}
    \includegraphics[width=\textwidth]{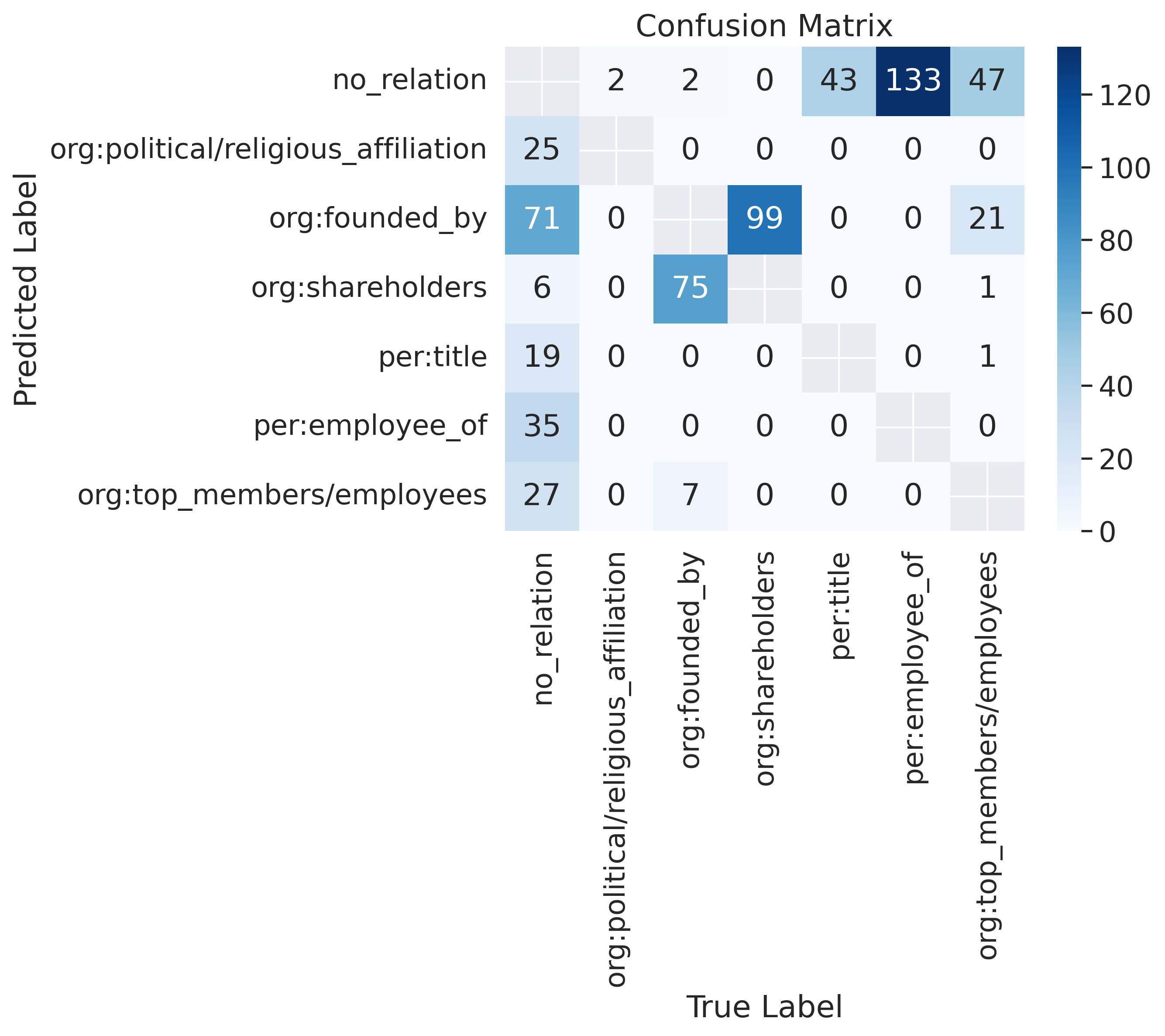}
    \caption{Confusion Matrix with Set Co-Expansion}
  \end{subfigure}
  \caption{Comparison of Confusion Matrices for 7 Selected Relations ReTACRED. We used the baseline PTR on the ReTACRED dataset to compare the difference between the original PTR and PTR with co-set expansion.}
  \label{fig:confusion_matrix_comparison}
\end{figure*}

The Table \ref{tb:results} provides insightful observations on the effectiveness of different models under varying conditions. The models were evaluated on three distinct datasets: TACREV, ReTACRED, and SEMEVAL.

The baseline performance, represented by the \textbf{original} category, establishes a fundamental measure for comparison. It is evident that the introduction of co-set expansion techniques brings noteworthy enhancements to the models' performance across most scenarios.
The comprehensive analysis of relation classification performance, as presented in Table 1, yields significant insights into the effectiveness of various models under diverse conditions. The models were meticulously evaluated across three distinct datasets: TACREV, ReTACRED, and SEMEVAL.

A prominent observation emerges even without constrastive scoring from the negative classes: the incorporation of co-set expansion techniques consistently benefits the performance of fine-tuned baselines across most scenarios. Integrating the co-set expansion without negative examples (\textbf{co w/o neg}) strategy already showcases remarkable improvements. This enhancement underscores the potency of harnessing contrastive scores derived from exemplars of contrastive classes. Also, this approach yields substantial gains in TACREV and ReTACRED datasets, suggesting its ability to effectively guide the models' learning process and improve their adaptability. This effect is particularly pronounced in the case of PTR, where fine-tuning with relational sub-prompts is employed. Leveraging exemplar expansion from seeded examples significantly enhances PTR's performance, resulting in a notable increase of over 2 percent on the TACREV and ReTACRED datasets.

Moreover, the influence of contrastive ranking and contrastive exemplar comparison becomes strikingly evident, particularly in the case of TACREV and ReTACRED. The utilization of contrastive information through co-set expansion (\textbf{co}) significantly boosts model performance. This approach facilitates the models' ability to disambiguate between contrasting relation categories, leading to substantial improvements in classification accuracy.

Figure \ref{fig:confusion_matrix_comparison} presents a deteiled comparison of confusion matrices for the 5 selected relations from . The left subfigure displays the original confusion matrix before the application of co-set expansion, while the right subfigure depicts the modified confusion matrix after co-set expansion. Notably, the co-set expansion technique not only enhances class distinctions but also reveals the underlying imbalance of errors within the confusion matrix. This visualization underscores the efficacy of co-set expansion in addressing class-related imbalances and improving overall classification performance.

Remarkably, the co-set expansion consistently demonstrates its potential in fine-tuning baselines, emphasizing its adaptability and generalizability. The co-set expansion methodology's success can be attributed to its ability to harness additional context from exemplars, which aids in distinguishing between contrasting relation categories.

\section{Conclusion}
\label{sec:conclusion}
In conclusion, our paper explores the effect of examplars for relation extraction, harnessing the power of language probing with analogous and contrastive examples from co-set expansion. We emphasize the significance of each step within the process and illustrate their combined impact on relation classification accuracy. This work not only advances the field of relation extraction but also offers valuable insights into how data augmentation and context-based tuning can collaboratively elevate the performance of information extraction systems.
\subsection{Limitations} However, our approach does have some limitations. Firstly, a significant challenge arises from the prevalence of errors that involve predicting no-relation as a well-defined relation, or vice versa. Mitigating these errors and addressing confusion between no-relation and specific relation categories remains a challenging aspect of our approach. Secondly, while our method seeds each relation category with representative exemplar pairs, the task of effectively selecting and seeding appropriate examples for each category remains a complex issue, especially considering that we currently rely on randomly sampling a limited number of exemplars. For example, in certain relations involving individuals and organizations, the exemplars might include pronouns, leading to potentially problematic seeds. Seeded examples from "per:siblings" could include instances like ("his", "Enzo"), ("her", "Lyle"), which might not provide accurate representation. Similarly, for the relation "per:parents," there could be unfavorable pairs like ("her", "Edda Mellas"), ("Carol Daniels", "his"), introducing noise into the seed set. Our approach involves seeding each relation category with representative exemplar pairs. However, the task of effectively selecting and seeding appropriate examples for each category remains intricate. This complexity is exacerbated by our current reliance on randomly sampling a limited number of exemplars, which might not fully capture the diversity and nuances of the relation patterns. This limitation underscores the challenge of striking a balance between comprehensiveness and the constraints of available data.

Additionally, the process of set co-expansion, while effective, can be computationally slow, particularly when dealing with large datasets. This computational overhead may impact the scalability of our approach, especially in scenarios where real-time or high-speed processing is required. Furthermore, the co-set expansion process introduces additional complexity to the training pipeline, potentially making it harder to interpret and analyze the resulting model's behavior.

Despite these limitations, our approach demonstrates promising results in improving relation classification accuracy and reducing confusion between classes. Future work could focus on addressing these limitations, exploring more efficient set co-expansion strategies, and refining the exemplar selection process to achieve even better performance and scalability.


\bibliographystyle{unsrt}  
\bibliography{references}

\end{document}